\documentclass[conference]{IEEEtran}
\IEEEoverridecommandlockouts
\usepackage{cite}
\usepackage{amsmath,amssymb,amsfonts}
\usepackage{algorithmic}
\usepackage{graphicx}
\usepackage{textcomp}
\usepackage{xcolor}
\usepackage{multirow}
\def\BibTeX{{\rm B\kern-.05em{\sc i\kern-.025em b}\kern-.08em
    T\kern-.1667em\lower.7ex\hbox{E}\kern-.125emX}}

\usepackage{tikz}
\usepackage{textcomp}
\usepackage{hyperref}
\usepackage{lipsum}

\newcommand\copyrighttext{%
  \footnotesize \textcopyright 2025 IEEE. Personal use of this material is permitted.
  Permission from IEEE must be obtained for all other uses, in any current or future
  media, including reprinting/republishing this material for advertising or promotional
  purposes, creating new collective works, for resale or redistribution to servers or
  lists, or reuse of any copyrighted component of this work in other works.}
\newcommand\copyrightnotice{%
\begin{tikzpicture}[remember picture,overlay]
\node[anchor=south,yshift=10pt] at (current page.south) {\fbox{\parbox{\dimexpr\textwidth-\fboxsep-\fboxrule\relax}{\copyrighttext}}};
\end{tikzpicture}%
}

\begin{document}

\title{Sparse Convolutional Recurrent Learning for Efficient Event-based Neuromorphic Object Detection\\
}

\author{\IEEEauthorblockN{Shenqi Wang$^{1}$, Yingfu Xu$^{2}$, Amirreza Yousefzadeh$^{3}$, Sherif Eissa$^{4}$, Henk Corporaal$^{4}$,\\Federico Corradi$^{4}$, Guangzhi Tang$^{5}$}
\IEEEauthorblockA{\textit{$^{1}$TU Delft, Delft, Netherlands $^{2}$imec, Eindhoven, Netherlands $^{3}$University of Twente, Enschede, Netherlands}\\
\textit{$^{4}$TU Eindhoven, Eindhoven, Netherlands $^{5}$Maastricht University, Maastricht, Netherlands}}
}

\maketitle
\copyrightnotice

\begin{abstract}

Leveraging the high temporal resolution and dynamic range, object detection with event cameras can enhance the performance and safety of automotive and robotics applications in real-world scenarios. However, processing sparse event data requires compute-intensive convolutional recurrent units, complicating their integration into resource-constrained edge applications. Here, we propose the Sparse Event-based Efficient Detector (SEED) for efficient event-based object detection on neuromorphic processors. We introduce sparse convolutional recurrent learning, which achieves over 92\% activation sparsity in recurrent processing, vastly reducing the cost for spatiotemporal reasoning on sparse event data. We validated our method on Prophesee's 1 Mpx and Gen1 event-based object detection datasets. Notably, SEED sets a new benchmark in computational efficiency for event-based object detection which requires long-term temporal learning. Compared to state-of-the-art methods, SEED significantly reduces synaptic operations while delivering higher or same-level mAP. Our hardware simulations showcase the critical role of SEED's hardware-aware design in achieving energy-efficient and low-latency neuromorphic processing.
\end{abstract}


\section{Introduction}
\label{intro}

Event cameras, known for their outstanding temporal resolution, wide dynamic range, and low data rate, are ideally suited for embedded and edge computing applications~\cite{gallego2020event}. Moreover, thanks to the advent of large-scale datasets~\cite{perot2020learning,de2020large}, it is possible to use event cameras for object detection in real-world applications such as automotive and robotics. In those applications, the properties of high dynamic range and low latency contribute to improved performance and safety. However, the inherently sparse data provided by event cameras demands a temporal learning capacity of the neural network suitable for accurate event-based object detection. So far, state-of-the-art event-based object detection solutions rely on resource-intensive convolutional recurrent units for effective spatiotemporal learning~\cite{perot2020learning,li2022asynchronous,gehrig2023recurrent}. Consequently, integrating these sophisticated solutions into resource-limited edge devices for low-power and low-latency applications poses a significant challenge.

Neuromorphic computing is well-suited for event-based vision, offering efficient solutions to process sparse event data streams from event cameras~\cite{xu2025event}. Event-based neuromorphic processors exploit input and activation sparsities of neural networks, significantly diminishing latency and power usage throughout inference~\cite{tang2023seneca}. Therefore, to fully harness the advantages of neuromorphic computing, it's crucial to employ neural networks that maintain sustainable activation sparsity. Spiking neural networks (SNNs), characterized by their sparse binary activations, are widely adopted in neuromorphic processors, providing energy-efficient solutions across diverse applications. Nevertheless, the complexity of object detection using event cameras poses a significant challenge for SNN-based solutions~\cite{cordone2022object}, hindering the broader adoption of efficient neuromorphic processors in tackling these complex tasks.

Interestingly, activation sparsity can be integrated into generic deep and recurrent neural networks by employing ReLU activation functions and thresholding operations. Notably, the EGRU (Event-based Gated Recurrent Unit) excels in sparse recurrent learning, balancing advanced temporal reasoning capabilities with reduced computational demands~\cite{subramoney2022efficient}. Recent advances in neuromorphic processors, including SpiNNaker 2~\cite{liu2018memory} and SENECA~\cite{tang2023seneca}, have paved the way for the efficient processing of generic neural networks with sparse activations, leveraging the support of graded spikes and flexible neural processing. This convergence between algorithm and hardware creates new opportunities for developing hardware-aware sparse recurrent learning solutions tailored for object detection with event cameras, showcasing their efficiency with neuromorphic computing.

In this paper, we propose the Sparse Event-based Efficient Detector (SEED) for high-performance and efficient object detection with event cameras \footnote{Code available here: https://github.com/ERNIS-LAB/SEED-Event-Object-Detection}. We integrated sparse convolutional recurrent (Conv-Rec) learning in SEED, significantly reducing the computational cost of Conv-Rec units. Additionally, SEED's fully convolutional architecture unlocks the potential to harness the hardware benefits of event-based neuromorphic processors with multi-core data-flow processing. We benchmarked our SEED on Prophesee's 1Mpx~\cite{perot2020learning} and Gen1~\cite{de2020large} datasets for event-based object detection. Our SEED considerably improves the object detection performance by 2.4$\times$ compared to the existing SNN-based neuromorphic solution~\cite{cordone2022object} while requiring only half of synaptic operations. Moreover, compared to state-of-the-art solutions that employ convolutional~\cite{perot2020learning} or transformer~\cite{gehrig2023recurrent} networks, our approach achieves higher or the same level of object detection performance with a 2$\times$ to 6$\times$ reduction in computational costs. We performed detailed hardware simulations employing the SENECA neuromorphic processor. The result indicates our hardware-aware design for event-based neuromorphic processing can directly translate to latency and energy consumption improvements on the hardware.

\section{Backgrounds and Related Works}
\label{background}

\subsection{Object Detection with Event Camera}

The event camera senses the brightness changes in the scene with high temporal precision~\cite{gallego2020event}. Therefore, compared with the frame-based camera, it is difficult to get meaningful object information from the instant outputs of the event camera when little changes emerge in a short time duration~\cite{kugele2023many}.

The state-of-the-art solutions rely on convolutional recurrent (Conv-Rec) units to learn spatial-temporal information. For instance, both RED~\cite{perot2020learning} and ASTMNet~\cite{li2022asynchronous} employ Conv-Rec layers for object detection with event frames. However, Conv-Rec units are costly in hardware as they require substantial amount of memory and synaptic operations. RVT~\cite{gehrig2023recurrent} attempts to resolve this problem by employing LSTM with 1x1 convolutions. Yet, the overall computational cost remains high, and the adoption of the vision transformer prevents it from being deployed on low-power event-based neuromorphic processors with data-flow processing. Recent event encoding methods achieve state-of-the-art object detection performance on event datasets without temporal information processing in the network~\cite{zubic2023chaos}. However, their performance across varied levels of instant event information remains undefined, potentially leading to high accuracy for objects with substantial event information but underperformance for those with low event data. Our SEED reduces the computation of Conv-Rec units by sparsifying the recurrent hidden activations. Furthermore, we compared the object detection performance across varied levels of instant event information and demonstrated the need for recurrent learning.

\subsection{Activation Sparsity and Sparse Recurrent Learning}

Neural networks designed for activation sparsity can harness the spatial-sparse data from event cameras, transforming the inherent sparsity of inputs into a benefit for computational efficiency. Spiking neural networks (SNN) leverages stateful spiking neurons and communicates through sparse binary spikes across layers. However, due to the complexity of the task and the limited capacity of binary activations, the current SNN solutions struggle to meet the state-of-the-art performance for event-based object detection~\cite{cordone2022object}. In addition to spiking neurons, activation sparsity also exists in deep neural networks with ReLU activation functions. Methods involving sparsity loss functions~\cite{georgiadis2019accelerating} and modified ReLU functions~\cite{kurtz2020inducing} have been introduced to enhance activation sparsity within deep neural networks. Nonetheless, these methods lack any form of temporal learning. The EGRU~\cite{subramoney2022efficient} adds a thresholding layer on the hidden state of the GRU~\cite{cho2014learning} to promote sparse recurrent computation. Yet, the effectiveness of the sparse recurrent learning in EGRU has not been proven in challenging tasks that demand robust spatiotemporal learning. Our SEED generalizes EGRU to sparse Conv-Rec learning and demonstrates the high performance of temporal learning on complex event-based object detection tasks.

\subsection{Event-based Neuromorphic Processor}

Event-based neuromorphic processors exploit the input sparsity from the event camera and the activation sparsity in neural networks. Additionally, the multi-core data-flow processing moves data directly to its consumed location, reducing the data-movement costs. An extensive range of neuromorphic processors only supports dedicated stateful spiking neurons~\cite{davies2018loihi,akopyan2015truenorth}. However, this approach stores all neural states in the on-chip memory, resulting in low mapping efficiency for convolutional layers~\cite{rueckauer2022nxtf}. Recent neuromorphic architecture designs promote flexible hybrid network that supports on stateful and non-stateful convolutional layers~\cite{pei2019towards,liu2018memory}. For example, SENECA processes non-stateful convolutional layers with significantly lower state memory requirements using specialized depth-first scheduling techniques~\cite{tang2023seneca}. Therefore, a hybrid network architecture with high activation sparsity is preferred for high-resolution event-based object detection on a neuromorphic processor with a limited form factor.

\section{Method}

\begin{figure*}[t]
\begin{center}
\centerline{\includegraphics[width=1.0\linewidth]{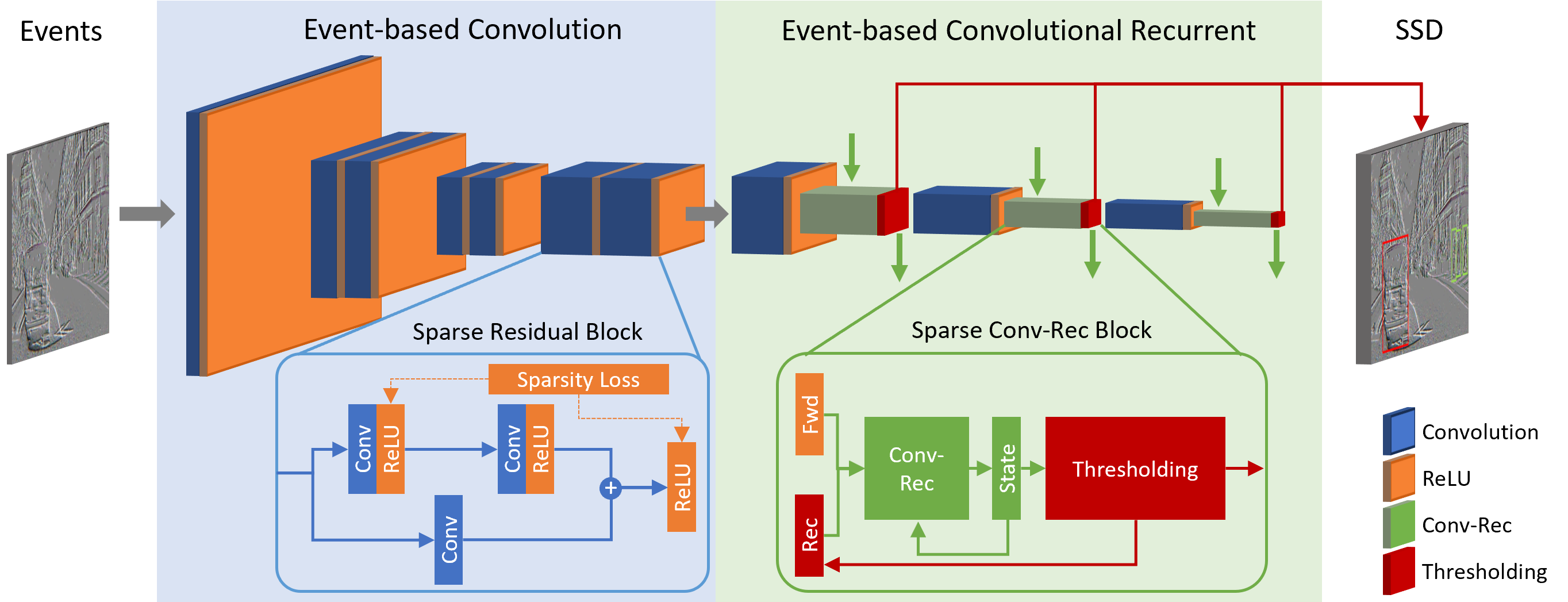}}
\caption{Sparse event-driven efficient detector (SEED) algorithm for event-based object detection. The event-based convolution module receives preprocessed event frames. The event-based Conv-Rec module receives events from the convolution module and the previous time step. The figure shows the processing of one event frame as an example. To emphasize the size of each layer, the figure accurately reflects the relative sizes of the layers' spatial and channel dimensions.}
\label{fig:network}
\end{center}
\vskip -0.2in
\end{figure*}

\subsection{Sparse Event-based Efficient Detector}

We proposed the SEED algorithm for solving event-based object detection. The SEED algorithm incorporates hardware-efficient event-based designs for neuromorphic processors with high-performance spatial-temporal learning. As shown in Figure~\ref{fig:network}, SEED's network architecture consists of three sub-modules: the event-based convolution module, the novel event-based convolutional recurrent (Conv-Rec) module, and the single-shot detector (SSD) head~\cite{liu2016ssd}.

The event-based convolution module downsamples the spatial dimension of the input using a series of sparse residual blocks. Employing the ReLU activation after convolution and, with a dedicated sparsity loss, sparsifies the layer's output, reducing computation in the subsequent layer. The convolution module can simultaneously extract spatial and short-term temporal information from the input with event preprocessing techniques that incorporate time into the channel dimension~\cite{sironi2018hats,rebecq2019high}. Our novel event-based Conv-Rec module extracts long-term temporal information from the scenes at different spatial scales. The sparse Conv-Rec block performs spatial-temporal feature extraction using very sparse events from the previous layer and the past step, ensuring low-cost event-based computations. The SSD head is used to predict the class and location of the object based on sparse multi-scale events from the Conv-Rec module.

\subsection{Event-based convolutional recurrent processing}

We extended and generalized the event-based recurrent processing in the EGRU~\cite{subramoney2022efficient} to the Conv-Rec module in SEED. Since the event-based Conv-Rec processing shares similar characteristics with EGRU, it retains the exact benefits of activation sparsity for inference and training, making it a perfect fit for neuromorphic processors that can exploit the activation sparsity. Our proposed event-based Conv-Rec processing generalizes to a broad spectrum of existing gated recurrent unit designs. We represent the convolutional operations for the gated recurrent unit in a basic form shown as follows,
\begin{equation}
\label{eq:gen-rec-unit}
    \mathbf{\alpha}^{(t)} = g(\mathbf{y}^{(t-1)}, \mathbf{x}^{(t)}),\ \mathbf{z}^{(t)} = f(\mathbf{y}^{(t-1)}, \mathbf{x}^{(t)})
\end{equation}
where $\mathbf{\alpha}$ is the forget gate computed by performing convolution operations on sparse input $\mathbf{x}$ at step $t$ from the previous layer and sparse hidden input $\mathbf{y}$ from the recurrent layer at step $t-1$, and $\mathbf{z}$ is the latent representation of inputs. The convolution operations only apply to these sparse inputs to maintain fully event-based processing in the recurrent unit. This basic form of convolutional recurrent processing applies to many existing gated recurrent unit designs, including GRU~\cite{cho2014learning}, LSTM~\cite{hochreiter1997long}, MGU~\cite{zhou2016minimal}, and MinimalRNN~\cite{chen2017minimalrnn}.

The hidden state $\mathbf{h}$ is the only dense state participating in the computation, which represents the memory in the recurrent unit. Since $\mathbf{h}$ is only involved in point-wise computations, the cost is neglectable compared to the convolutional operations defined in Equation~\ref{eq:gen-rec-unit}. We define a basic form of hidden state update at every step as follows,
\begin{equation}
\label{eq:state-update}
    \mathbf{h}^{(t)} = \mathbf{\alpha}^{(t)} \odot \mathbf{h}^{(t-1)} + (1-\mathbf{\alpha}^{(t)})\odot \mathbf{z}^{(t)} - \mathbf{s}^{(t-1)} \odot \mathbf{V}_{th}
\end{equation}
where $\mathbf{s}^{(t-1)} \odot \mathbf{V}_{th}$ is a soft reset of hidden state inspired by spiking neurons~\cite{guo2022reducing} to help forgetting, $\mathbf{s}^{(t-1)}$ is the binary hidden events at step $t-1$, and $\mathbf{V}_{th}$ is the learnable threshold for each neuron. The Conv-Rec unit generates events using a Heaviside step function $H(\cdot)$ on the hidden state at every step,
\begin{equation}
\label{eq:spike-gen}
    \mathbf{s}^{(t)} = H(\mathbf{h}^{(t)} - \mathbf{V}_{th})
\end{equation}
\begin{equation}
\label{eq:event-gen}
    \mathbf{y}^{(t)} = \mathbf{h}^{(t)} \odot \mathbf{s}^{(t)}
\end{equation}
where the hidden output $y$ is used as the hidden input at step $t+1$ and the input to the subsequent layer at step $t$. We used a surrogates gradient to estimate the backpropagated gradient of $H(\cdot)$ and train the threshold $\mathbf{V}_{th}$ of each neuron using the same methods presented in~\cite{subramoney2022efficient}. After training, each neuron in the hidden state has a unique threshold ranging from zero to one.

\subsection{Sparse residual block and activation sparsity loss}

The ReLU activation functions in the residual block induce sparsity in activation maps, and reduce synaptic operations in the subsequent convolutional layer. However, activation sparsity in standard residual blocks is typically low, within a range between 20\% to 50\%~\cite{georgiadis2019accelerating,kurtz2020inducing}. This causes the event-based convolution module to dominate the computational cost of SEED, mainly when extremely high sparsities are achieved in the Conv-Rec module. Therefore, to further reduce the computational cost of SEED, we applied the L1 norm on the activation maps within the residual blocks as an activation sparsity loss function $L_{sparse}$~\cite{georgiadis2019accelerating}, defined as follows,
\begin{equation}
\label{eq:sparse_loss}
    L_{sparse}=\beta_{sparse}\frac{1}{NT}\sum_{n=1}^{N}\sum_{t=1}^{T}\sum_{l=1}^{L} ||\mathbf{a}^{(n)(t)(l)}||_{1}
\end{equation}
where $\beta_{sparse}$ is the coefficient of the loss, $n$ is the index of the training sample, $N$ is the mini-batch size, $t$ is the index of the step within a training sample, $T$ is the overall steps of a training sample, $l$ is the index of the layer, $L$ is the overall number of layers, and $\mathbf{a}$ denotes the activation maps, which refers to all ReLU outputs.

As shown in Figure~\ref{fig:network}, the sparsity loss applies to the selected activation maps in the first convolutional layer and the subsequent sparse residual blocks, which can directly reduce the synaptic operations in the succeeding layer. Adhering to the widely accepted methodology outlined in~\cite{georgiadis2019accelerating,kurtz2020inducing}, we finetuned a sparsity-aware SEED by combining the sparsity loss with the object detection loss, building upon a pre-trained model without sparsity loss. The finetuning approach normally performs better than training with sparsity loss from scratch. Additionally, we performed a thorough hyperparameter search for $\beta_{sparse}$ to determine the optimal sparsity loss factor, which balances the object detection performance and computational cost.

\section{Experiment and Results}

\begin{table*}[t]
\caption{Comparing SEED with state-of-the-art object detection approaches for event-based vision}
  \label{tb:main-result}
  \vskip 0.15in
  \begin{center}
\begin{tabular}{ccccccc}
\hline
 \multicolumn{2}{c|}{} & \multicolumn{2}{c|}{1 Mpx} & \multicolumn{2}{c|}{Gen1} &  \\ 
\hline
Method & \multicolumn{1}{c|}{Network} & mAP  & \multicolumn{1}{c|}{GSOp} & mAP  & \multicolumn{1}{c|}{GSOp} & Param(M)\\
\hline
ASTMNet~\cite{li2022asynchronous} & \multicolumn{1}{c|}{TACN+ConvRec+SSD} & 48.3  & \multicolumn{1}{c|}{-} & 46.7  & \multicolumn{1}{c|}{-} & $>$39.6\\
SNN~\cite{cordone2022object} & \multicolumn{1}{c|}{Spiking DenseNet+SSD} & -  & \multicolumn{1}{c|}{-} & 18.9  & \multicolumn{1}{c|}{2.33} & 8.2\\
SpikeYOLO~\cite{10.1007/978-3-031-73411-3_15} & \multicolumn{1}{c|}{Spiking+YOLOv8} & - & \multicolumn{1}{c|}{-} & 40.4 & \multicolumn{1}{c|}{14.3} & 23.1\\
RED~\cite{perot2020learning} & \multicolumn{1}{c|}{SENet+ConvLSTM+SSD} & 43.0  & \multicolumn{1}{c|}{26.1} & 40.0 & \multicolumn{1}{c|}{8.26} & 24.1 \\
RVT-B~\cite{gehrig2023recurrent} & \multicolumn{1}{c|}{MaxViT+LSTM+YOLOX} & 47.4 & \multicolumn{1}{c|}{15.6} & 47.2 & \multicolumn{1}{c|}{5.05} & 18.5\\
RVT-S~\cite{gehrig2023recurrent} & \multicolumn{1}{c|}{MaxViT+LSTM+YOLOX} & 44.1 & \multicolumn{1}{c|}{8.69} & 46.5 & \multicolumn{1}{c|}{2.78} & 9.9\\
RVT-T~\cite{gehrig2023recurrent} & \multicolumn{1}{c|}{MaxViT+LSTM+YOLOX} & 41.5 & \multicolumn{1}{c|}{3.87} & 44.1 & \multicolumn{1}{c|}{1.29} & 4.4\\
\hline
\textbf{SEED-256 (Ours)} & \multicolumn{1}{c|}{ECNN+EConvGRU+SSD} & 44.9  & \multicolumn{1}{c|}{3.83} & 45.3 & \multicolumn{1}{c|}{1.32} & 13.9 \\
\textbf{SEED-128 (Ours)} & \multicolumn{1}{c|}{ECNN+EConvGRU+SSD} & 44.1  & \multicolumn{1}{c|}{2.75} & 44.5 & \multicolumn{1}{c|}{0.99} & 4.8 \\
\hline
\end{tabular}
  \end{center}
  \vskip -0.1in
\end{table*}

Our experiments are designed to achieve three key objectives. Firstly, we showcased SEED by benchmarking its performance against state-of-the-art approaches, as well as testing its compatibility with various types of recurrent units. Secondly, we highlighted the need for recurrent learning on event-based object detection and demonstrated the effectiveness of sparse recurrent learning. Lastly, we analyzed the influence of high-sparsity event-based processing within the convolutional modules. We evaluated our method on the 1 Mpx~\cite{perot2020learning} and Gen1~\cite{de2020large} datasets following the standard approach presented in~\cite{perot2020learning} and reported the widely used COCO mAP~\cite{lin2014microsoft}. We employed GRU in event-based Conv-Rec modules except in Section~\ref{sec:gen-rnn}. We performed two-stage training for SEED. The best model on the validation set from the first state is used for sparsity-aware fine-tuning in the second stage. The best model on the validation set from the second stage is applied to the test set to report the final mAP. Additionally, we recorded the average activation sparsity of each layer in SEED when performing inference on the test set to compute its effective synaptic operations. 

\subsection{Experiment setup}
The proposed neural network SEED includes two parts: the backbone and the SSD detection head. The architecture of the backbone is shown in Table~\ref{tab:Backbone Architecture of SEED-256}:
\begin{table}[htbp]
\caption{Backbone Architecture of SEED-256}
    \label{tab:Backbone Architecture of SEED-256}
    \vskip 0.15in
  \begin{center}
    \begin{tabular}{c c c c}
    \hline
    Layer & Channels out & Dimension out(1Mpx) & Stride \\\hline
    ConvBNReLU & 32 & [320, 180] & 2 \\
    Residual Connection & 64 & [160, 90] & 2 \\
    Residual Connection & 64 & [160, 90] & 1 \\
    Residual Connection & 128 & [160, 90] & 1 \\
    EventConvRNN & 256 & [80, 45] & 2 \\
    EventConvRNN & 256 & [40, 23] & 2 \\
    EventConvRNN & 256 & [20, 12] & 2 \\\hline
    
    \end{tabular}
    \end{center}
  \vskip -0.1in
\end{table}

In our model, we employed distinct initialization strategies for different components. All convolutional kernel parameters were initialized using a uniform distribution method. For the Event-based Conv-Rec layers, all trainable threshold parameters were initialized using a normal distribution with zero mean and standard deviation of $\sqrt{2}$.

All models are trained with full precision, using ADAM optimizer. We divided the training into two stages, each consisting of 35 epochs.  For both stages, we utilized the OneCycle scheduler to adjust the learning rate dynamically. For both 1Mpx and Gen1 datasets, the maximum learning rate is set at 2.5e-4 for stage 1 and 1e-4 for stage 2. On the 1Mpx dataset,  we train our models with a batch size of 4, sequence length of 12. On the Gen1 dataset,  we train our models with a batch size of 5, sequence length of 24. 

During stage 1, the focus was solely on enhancing the model's object detection capabilities, which involved training with only the object detection loss as in~\cite{perot2020learning}. This loss follows the SSD formulation and consists of two components: a focal classification loss applied to all anchors, and a Smooth L1 regression loss computed only for positive anchors, both normalized by the number of positive samples. In stage 2, we introduced the $L_{sparse}$ activation sparsity loss term into the training process, with its coefficient $\beta_{sparse}$ set at 0.04.

Following the prior work~\cite{perot2020learning}, we implemented a strategy for bounding box selection tailored to each dataset's characteristics during the training and evaluation stage. For the 1Mpx dataset, we excluded bounding boxes with diagonals less than 60 pixels and sides less than 20 pixels. In contrast, for the Gen1 dataset, bounding boxes with diagonals smaller than 30 pixels and sides less than 10 pixels are removed.

\subsection{Benchmarking SEED against the state-of-the-arts}

We benchmarked SEED against state-of-the-art object detection methods for event-based vision in Table~\ref{tb:main-result}. Given the crucial role of long-term memory in event-based object detection, we focused on methods featuring explicit network designs for temporal learning in the benchmarking. Beyond traditional metrics like object detection performance (mAP) and network size (Parameters), our analysis also contained synaptic operations (GSOp) to offer a comprehensive view of the hardware expenses associated with each network inference. In our analysis, the synaptic operation is denoted as ACC (Accumulate) within the SNN, and as MAC (Multiple-and-accumulate) across the other network architectures. For RED and RVT, we computed their GSOp using the open-sourced implementations. Additionally, we considered effective synaptic operations for SNN, RED, and our SEED by skipping the zero values when activation sparsity is available within the layers for event-based depth-first convolution performing on the SENECA neuromorphic processor.

Table~\ref{tb:main-result} showcases two variants of SEED utilizing different channel dimensions within the event-based ConvGRU layers: one is a high-performance standard model (SEED-256), and the other, an efficient yet marginally less performant alternative (SEED-128). Our method, especially SEED-128, significantly outperforms the existing neuromorphic approach that employs spiking neurons~\cite{10.1007/978-3-031-73411-3_15}, achieving 5.1 improvements in mAP while requiring 9\% of synaptic operations and 56\% of parameters. Moreover, our method sets itself apart from RED~\cite{perot2020learning} through its hardware-aware design tailored for event-based neuromorphic processing. Compared to RED, the SEED-256 increases mAP (by 1.9 in the 1Mpx dataset and 5.3 in the Gen1 dataset) with a more than 6$\times$ synaptic operation reduction. This result, in conjunction with the hardware simulation analyses detailed in Section~\ref{sec:hardware-sim}, underscores the critical role of our hardware-aware design in achieving energy-efficient and low-latency neuromorphic processing.

Furthermore, our fully convolutional SEED network beats the RVT-S~\cite{gehrig2023recurrent} by reducing more than 2$\times$ on GSOp while achieving similar mAP. The transformer-based RVT comprises efficient attention mechanisms from MaxViT and LSTM with 1$\times$1 convolutions, which is inherently more efficient than fully convolutional designs with 3$\times$3 Conv-Rec layers. Here, we demonstrated that event-based convolution, characterized by high activation sparsity, outperforms transformer-based architectures in terms of efficiency. Additionally, the fully convolutional design enables deployments on neuromorphic processors with data-flow processing, translating the theoretical efficiency into tangible reductions in energy consumption and inference latency on hardware.

\subsection{Generalizing to variations of recurrent unit}
\label{sec:gen-rnn}

\begin{figure*}[t]
\begin{center}
\centerline{\includegraphics[width=1.0\linewidth]{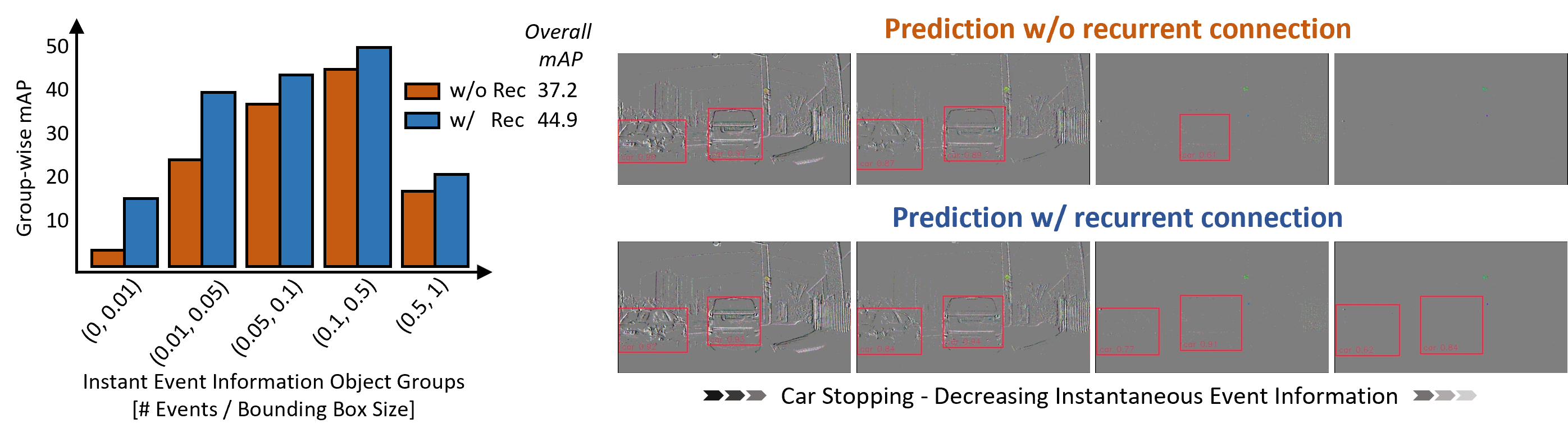}}
\caption{Comparing the object detection performance of SEED-256 and its recurrent-free ablation on the 1Mpx dataset. Left - Overall and group-wise mAP of the two models. The group-wise mAP represents the model's object detection capability on ground truth bounding boxes with different levels of instant event information. Right - Visualization of the event frames capturing a car stopping scene and the predicted bounding boxes by the two models. The text in a bounding box describes the class and confidence of the prediction.}
\label{fig:recurrent}
\end{center}
\vskip -0.2in
\end{figure*}

\begin{table}[t]
  \caption{Comparing SEED with different recurrent units}
  \label{tb:recurrent-units}
  \vskip 0.15in
  \begin{center}
  \begin{tabular}{llll}
    \hline
    Method     & mAP & GSOp & Param(M)\\
    \hline
    SEED+GRU & 44.9 & 3.83 & 13.9 \\
    SEED+MinimalRNN & 43.7 & 3.93 & 7.9\\
    SEED+MGU & 44.7 & 3.45 & 10.7 \\
    SEED+LSTM & 40.1 & 4.16 & 15.9 \\
    \hline
  \end{tabular}
  \end{center}
  \vskip -0.1in
\end{table}

We extended the event-based Conv-Rec processing to various popular recurrent units, demonstrating the broad applicability of our proposed method. Specifically, we generalized the approach to LSTM~\cite{hochreiter1997long}, MinimalRNN~\cite{chen2017minimalrnn}, and MGU~\cite{zhou2016minimal}. Since LSTM contains two states, we applied Equation~\ref{eq:state-update} on the cell state for state update and Equation~\ref{eq:event-gen} on the hidden state for event generation. MinimalRNN and MGU are simplified gated recurrent unit designs. Both designs contain a single hidden state, which we can directly use for event generation.

We validated SEED-256 with different recurrent units on the 1 Mpx dataset~\cite{perot2020learning} and compared the results in Table~\ref{tb:recurrent-units}. Deploying event-based Conv-Rec processing on all recurrent units delivers competitive performance, with consistently less than 7\% event density generated by recurrent layers. Notably, MGU performs similarly to GRU while requiring fewer parameters and synaptic operations. However, although LSTM uses more parameters, we observed performance reduction compared to other recurrent units. This shows that event-based processing may limit the temporal learning capacity due to either the gate complexity or the dual states in the LSTM.

\subsection{Effectiveness of sparse recurrent learning}

Since the event camera only captures brightness changes, low relative movements between objects and the camera activate very few events within a short time window. Therefore, detecting these objects with low instant event information requires utilizing temporal information from the past. We investigated the effectiveness of SEED's sparse recurrent learning on detecting objects with low instant event information by performing an ablation study on recurrent units. The study comprises two steps. Firstly, we trained a recurrent-free variation of SEED-256 that zeroed out all temporal information in the event-based recurrent unit, including recurrent events and hidden states. Secondly, we compared SEED and its recurrent-free variation on detecting objects with different amounts of instant event information in the input. As shown in Figure~\ref{fig:recurrent}, we divided the ground truth bounding boxes into 5 groups based on the instant event information. Then, we computed the group-wise mAP of the models by splitting predicted bounding boxes into their corresponding groups and comparing them with the ground truth bounding boxes in the group.

Compared to the recurrent-free model, our SEED performs significantly better object detection in the 1 Mpx dataset, despite the recurrent hidden events exhibiting a sparsity exceeding 95\%. Particularly, the group-wise mAP in Figure~\ref{fig:recurrent} shows SEED gains much higher performance improvements than the recurrent-free model on groups with low instant event information. These results prove the effectiveness of sparse recurrent learning in SEED in utilizing temporal information to accurately detect objects that cannot be detected using instant information. To visualize the group-wise performance differences between the two models, we selected a car-stopping scene in which the number of events captured by the event camera gradually decreases over time. The right-most event frame in Figure~\ref{fig:recurrent} has nearly zero events on the two cars in the front when the car with the camera stops. Therefore, recurrent learning in SEED is required to use information from the event frames in the past to detect the two cars.

\subsection{Effectiveness of sparsity-aware training}

To achieve effective sparsity-aware training, the model must simultaneously observe substantial sparsity improvement in the activation maps and minimal object detection performance drop. We compared the finetuned sparsity-aware SEED with the pre-trained model before applying the sparsity loss in Equation~\ref{eq:sparse_loss}. Table~\ref{tb:sparse-training} presents the comparison, detailing both the variations in total synaptic operations and shifts in object detection performance, transitioning from the pre-trained model to the sparsity-aware SEED. The result shows that sparsity-aware training significantly reduces the number of synaptic operations in SEED while maintaining or even improving the object detection performance. The accuracy improvements have also been observed in other work performing activation sparsity-aware training~\cite{georgiadis2019accelerating}, which can potentially be due to the regularization effect of the sparsity loss for less over-fitting and better generalization.

\begin{table}[h]
\caption{Sparsity-aware training improvements}
  \label{tb:sparse-training}
  \vskip 0.15in
  \begin{center}
\begin{tabular}{ccccc}
\hline
 & \multicolumn{2}{c}{1 Mpx} & \multicolumn{2}{c}{Gen1}\\ 
\hline
Method & GSOp  & \multicolumn{1}{c|}{mAP} & GSOp  & mAP\\
\hline
\textbf{SEED-256} & -1.86(-33\%)  & \multicolumn{1}{c|}{+0} & -0.68(-34\%) & +0.9 \\
\textbf{SEED-128} & -2.04(-43\%)  & \multicolumn{1}{c|}{+0.2} & -0.40(-29\%) & +0.6 \\
\hline
\end{tabular}
\end{center}
  \vskip -0.1in
\end{table}

\section{Hardware Simulation on Event-based Neuromorphic Processor}
\label{sec:hardware-sim}

\begin{figure*}[t]
\begin{center}
\centerline{\includegraphics[width=1.0\linewidth]{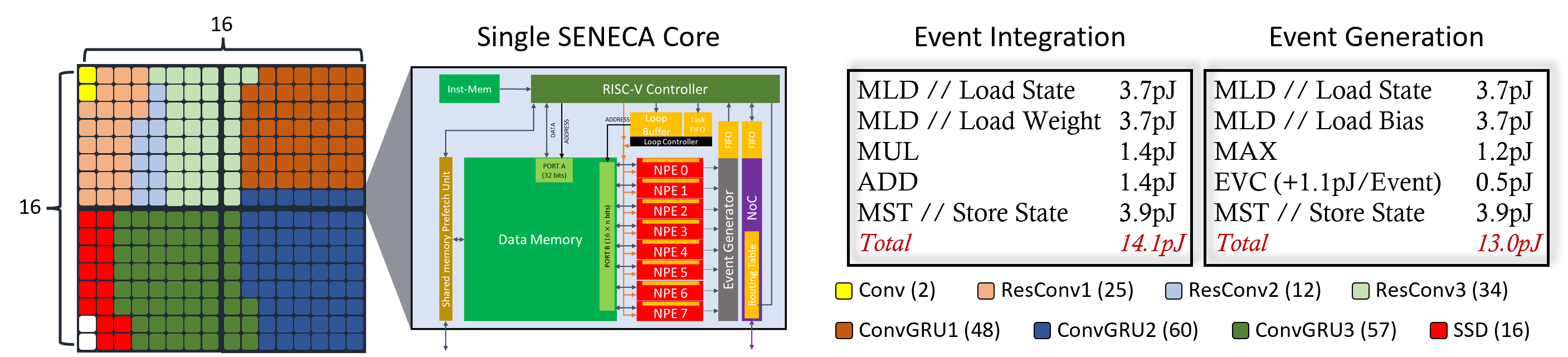}}
\caption{Hardware simulation mapping of SEED-256 on a 256-core SENECA neuromorphic processor. Each SENECA core contains an on-chip memory, 8 NPEs, and other controller and accelerators for event control, pre-processing, post-processing, and communications. The micro-kernels show detailed instructions and corresponding actual hardware energy costs for event integration and generation.}
\label{fig:hardware}
\end{center}
\vskip -0.2in
\end{figure*}

We showcase the benefits of SEED's hardware-aware design by conducting hardware simulations employing an event-based digital neuromorphic processor. Given its representative event-based and data-flow neural computing paradigm, we selected SENECA~\cite{tang2023seneca} as the reference hardware to ensure that our conclusions are applicable across a broad spectrum. The simulation comprises analytical studies of network activities based on actual hardware measurements~\cite{tang2023open}, demonstrating that SEED's hardware-aware design results in advantageous energy consumption, latency, and silicon area.

\subsection{Hardware simulation employing SENECA}

\begin{table*}
  \caption{Comparing hardware simulation results of SEED and different variations of the event-based object detector}
  \label{tb:hw-mapping-result}
  \vskip 0.15in
  \begin{center}
  \begin{tabular}{cccccccccc}
    \hline
     & Recurrent & Sparse & Sparsity & Recurrent & Energy & Latency & Memory & Cores & GSOp\\
    Method & Dimension & Recurrent & Loss & Unit & (mJ)   & (ms)    & (Mb) & & \\
    \hline
    \multirow{2}{*}{\textbf{SEED}} & 128 & Yes & Yes & 3$\times$ GRU & 39.3 & 21.6 & 90.8 & 236 & 2.75 \\
    & 256 & Yes & Yes & 3$\times$ GRU & 54.7 & 44.9 & 245.1 & 254 & 3.83 \\
    \hline
    \multirow{3}{*}{\shortstack{SEED \\ (w/ ablations)}} & 128 & Yes & No & 3$\times$ GRU & 68.1 & 44.0 & 90.8 & 256 & 4.79 \\
    & 256 & Yes & No & 3$\times$ GRU & 80.9 & 81.43 & 245.1 & 254 & 5.69 \\
    & 256 & No & No & 3$\times$ GRU & 284.3 & 240.1 & 226.1 & 234 & 20.1 \\
    \hline
    \multirow{2}{*}{\shortstack{RED \\ (w/o SE)}} & 256 & No & No & 5$\times$ LSTM & 368.6 & 1077.3 & 404.1 & 343 & 26.1 \\
    & 256 & No & No & 5$\times$ LSTM & 368.6 & 40.7 & 404.1 & 1446 & 26.1 \\
    \hline
  \end{tabular}
  \end{center}
  \vskip -0.1in
\end{table*}

SENECA is a digital neuromorphic architecture specializing in processing neural networks with high activation sparsity. The architecture design supports scaling up to 256 cores in a single chip, with each core comprising 2 Megabits of on-chip memory and 8 neuron processing elements (NPEs). The NPE accelerates a rich neuron processing instruction set. A neural network runs on SENECA by sequentially executing multi-instruction micro-kernels on the NPEs. For example, Figure~\ref{fig:hardware} shows two micro-kernels conducting event integration and generation for convolutional layers with the ReLU activation. SENECA performs event-based depth-first convolution, which exploits the activation sparsity and executes data-flow processing to realize inter-layer parallelism~\cite{xu2024optimizing}. Additionally, the depth-first processing considerably reduces the state memory costs of non-stateful convolutional layers.

The hardware simulation of event-based neural network processing consists of two elements. First, the network is mapped on a 256-core neuromorphic processor. The mapper performs channel-wise mapping, which simultaneously distributes each layer's memory requirements and computation to multiple cores. The optimized mapping limits the memory usage of each core within the on-chip memory capacity and minimizes the computational load of the busiest core. Second, we benchmarked the energy cost and latency of the network inference. We constructed micro-kernels for computations required by the event-based network inference. The micro-kernels execute neural computations and memory accesses, providing a complete picture of hardware costs. Since the design frequency of a SENECA core is 500 MHz, each instruction in the micro-kernel requires 2 ns to execute. For event-based depth-first convolution, the layer with the highest computational load dominates the latency of the network inference. Therefore, we took the busiest core's latency as the network's latency in the simulation. We obtained the energy costs of instructions from actual hardware measurements~\cite{tang2023open}. We added the energy cost of all instructions executed in the inference to compute the overall energy consumption. In both mapping and benchmarking, we determine the average required computation of a layer by the average activation density of its inputs over the test data of the 1Mpx dataset.

\subsection{Advantages of hardware-aware SEED design}

We performed hardware simulation for SEED and different variations of the event-based object detector to demonstrate the advantages of SEED's hardware-aware design. Table~\ref{tb:hw-mapping-result} compares the hardware costs of different networks. The simulation shows synaptic operations dominate the energy consumption since the number of event integrations is orders of magnitude higher than other kinds of operations (e.g., event generation and element-wise matrix computation in recurrent units). On the contrary, the latency is impacted by various factors, such as the flexibility of the mapper considering memory constraints and the distribution of layerwise computational load.

We simulated SEED with varying sizes to quantify the hardware cost improvements when reducing the recurrent dimension. Interestingly, SEED-128 reduces latency by 2$\times$, surpassing the 1.4$\times$ overall network synaptic operation reduction. This is due to the lower memory cost of SEED-128 compared to SEED-256, which gives the mapper additional resources to distribute high-latency layers in multiple cores. 

Moreover, we simulated SEED with ablations on sparsity. The results show that SEED's sparsity-aware design delivers a 5$\times$ reduction in energy and latency compare to the model without event-based Conv-Rec units and activation sparsity loss. Notably, the proposed event-based Conv-Rec unit contributes over 80\% to the energy and latency reduction of the comparison, claiming to be the primary source of hardware-cost improvements.

Lastly, we simulated a data-flow-friendly version of RED~\cite{perot2020learning} by replacing the Squeeze-and-Excitation blocks~\cite{hu2018squeeze} with sparse residual blocks. Due to the high memory costs of ConvLSTM layers, our mapper failed to map the RED on a 256-core processor. Therefore, we provided two different mappings of RED in Table~\ref{tb:hw-mapping-result}. First, we mapped RED on the minimal number of cores required based on on-chip memory constraints. As expected, this mapping results in extremely high latency. Then, the mapper reduces the latency of RED with unlimited number of cores. The RED achieves similar latency to SEED-256 while using 5.7$\times$ more cores. Consequently, since the hardware area correlates with the chip manufacturing cost, SEED is significantly cheaper than RED for deploying on neuromorphic processors.

\section{Discussion and Conclusion}

This paper presents SEED, a computationally efficient neuromorphic detector for event-based object detection. The method effectively addresses the challenge of resource-intensive convolutional recurrent processing by adopting innovative sparse convolutional recurrent learning. Consequently, we set a new benchmark in recurrent learning for efficient object detection using event cameras.

With neuromorphic computing embracing generic neural networks~\cite{pei2019towards,liu2018memory,schuman2022opportunities}, our method is positioned to substitute SNNs as the dominant method for event-based vision on neuromorphic processors, when high performance is required with efficient computing. Employing comprehensive hardware simulation, our study explores the efficiency gain of SEED's hardware-aware design. This paves the way for versatile neuromorphic solutions for real-world edge applications.

Our sparse convolutional recurrent processing excels in long-term temporal reasoning. Its potential to integrate with recent event preprocessing methods promises even higher performance gains~\cite{zubic2023chaos,peng2023better}. The temporal learning in our approach unlocks the possibility for efficient ultra-low-latency object detection with exceedingly short event time bins, which will be critical for high-speed automotive and robotics applications.

\bibliography{reference}
\bibliographystyle{IEEEtran}

\end{document}